\pdfoutput=1

\documentclass[11pt]{article}

\usepackage{acl}

\usepackage{times}
\usepackage{latexsym}

\usepackage{CJKutf8}
\usepackage{makecell}
\usepackage[ruled]{algorithm2e}
\usepackage{amsthm}
\usepackage{amsmath}
\usepackage{amsfonts}
\usepackage{multirow}
\usepackage{enumitem}
\usepackage{array}
\usepackage{mathrsfs}
\usepackage{stmaryrd} 
\usepackage{subcaption}
\usepackage{balance}
\usepackage{graphicx}
\usepackage{amsfonts}
\usepackage{url}
\usepackage{xspace}
\usepackage{upgreek}
\usepackage{color,xcolor}
\usepackage{amsmath}
\usepackage{makecell}
\usepackage{multirow}
\usepackage{array}
\usepackage{booktabs}
\usepackage{amssymb}
\usepackage{amsmath}
\usepackage{amsfonts}
\usepackage{multirow}
\usepackage{booktabs}
\usepackage{xspace}

\usepackage[T1]{fontenc}

\newcommand{\stitle}[1]{\vspace{1ex} \noindent{\bf #1}}

\newcommand{\modelname}{\textsc{MHP}\xspace}

\usepackage[utf8]{inputenc}

\usepackage{microtype}
\usepackage{booktabs}
\usepackage{multirow}

\title{\mbox{Dangling-Aware Entity Alignment with Mixed High-Order Proximities}}

\author{\textbf{Juncheng Liu}$^{1}$ \quad
  \textbf{Zequn Sun}$^{2}$ \quad
  \textbf{Bryan Hooi}$^{1}$ \quad
  \textbf{Yiwei Wang}$^{1}$ \quad
  \\
  \textbf{Dayiheng Liu}$^{3}$ \quad
  \textbf{Baosong Yang}$^{3}$ \quad
  \textbf{Xiaokui Xiao}$^{1}$ \quad
  \textbf{Muhao Chen}$^{4}$ 
  \\
  $^{1}$National University of Singapore \quad
  $^{2}$Nanjing University \\
  $^{3}$Alibaba Group \quad
  $^{4}$University of Southern California \\
  \texttt{juncheng.liu@u.nus.edu}, \texttt{zqsun.nju@gmail.com} \\
  \texttt{muhaoche@usc.edu}
  }

\begin{document}
\maketitle
\begin{abstract}
We study dangling-aware entity alignment in knowledge graphs (KGs), which is an underexplored but important problem. 
As different KGs are naturally constructed by different sets of entities, a KG commonly contains some dangling entities that cannot find counterparts in other KGs. 
Therefore, dangling-aware entity alignment is more realistic than the conventional entity alignment where prior studies simply ignore dangling entities.
We propose a framework using mixed high-order proximities on dangling-aware entity alignment.
Our framework utilizes both the local high-order proximity in a nearest neighbor subgraph and the global high-order proximity in an embedding space for both dangling detection and entity alignment. 
Extensive experiments with two evaluation settings shows that our 
framework more precisely detects dangling entities, and better aligns matchable entities.
Further investigations demonstrate that our framework can mitigate the hubness problem on dangling-aware entity alignment. 

\end{abstract}

\section{Introduction}


Knowledge graphs (KGs)
have become 
the backbone of many intelligent applications \citep{ji2021survey}. 
In spite of their importance, many KGs are 
independently created without considering the
interrelated and interchangeable nature of 
individually created 
knowledge \citep{chen2020multilingual}. 
To allow complementary
knowledge to be automatically combined and migrated across individual KGs, entity alignment seeks to identify equivalent entities in distinct KGs \cite{sun-etal-2020-knowledge}.
Recent literature has focused on learning embedding representations of multiple KGs where 
identical entities are aligned based on their embedding similarity  \citep{chen_MTransE,cao-etal-2019-multi, Fey2020Deep, sun-etal-2020-knowledge,liu2021visual}.

\begin{figure}[t]
    \centering
  \begin{minipage}[t]{0.5\linewidth}
    \subcaptionbox{\label{fig:local_high_order}}
    {\includegraphics[height=5.0cm]{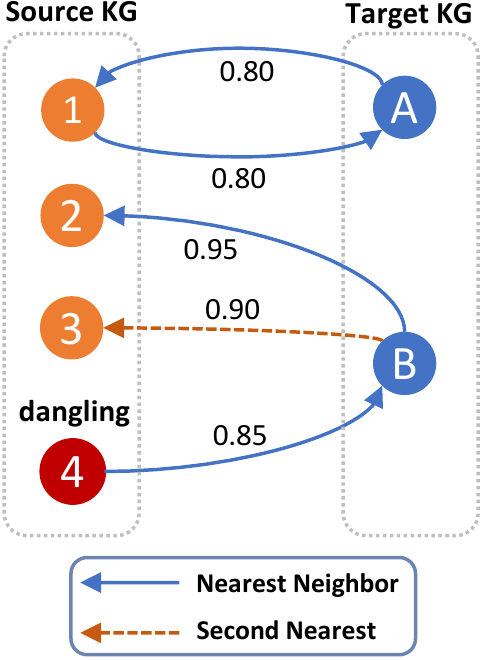}}
  \end{minipage}%
  \begin{minipage}[b]{0.5\linewidth}
    \centering
    \subcaptionbox{\label{fig:global_dist_a}}
    {\includegraphics[scale=0.5]{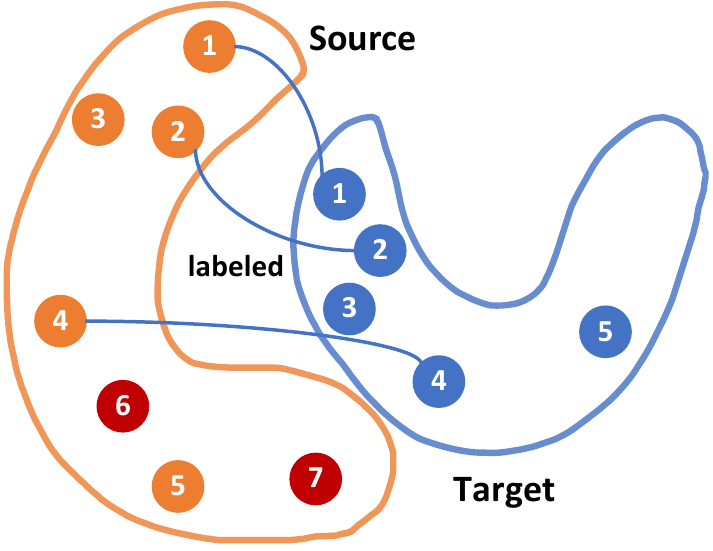}} \\
    \subcaptionbox{\label{fig:global_dist_b}}
    {\includegraphics[scale=0.55]{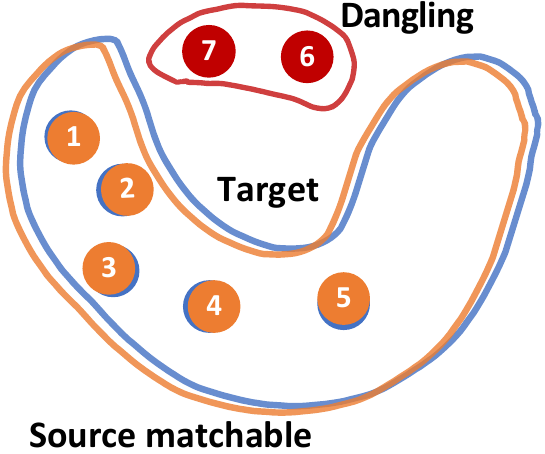}}
  \end{minipage}
    \vspace{-0.5em}
    \caption{Examples for mixed high-order proximities. 
    (a) 
    Nearest neighbor (NN) subgraph where entities connect to NNs in the other KG using embedding similarities.
    4 and its nearest neighbor B have 0.85 similarity. B prefers 2 and 3 with higher similarities. 1 and A are mutual nearest neighbors. 
    (b) Labeled alignments and dangling entities. (c) Aligning matchable source and target distributions rather than only labeled alignments. }  
    \vspace{-1em}
\end{figure}


Aside from the surge of research effort on entity alignment \cite{zeng2021comprehensive}, an unresolved but important challenge that existing methods face is
the \textit{dangling entity} problem.
Dangling entities 
are those unique entities in a KG that cannot find counterparts in another KG.
Considering that individually created KGs are unlikely to share the same set of entities, identifying dangling entities is undoubtedly an indispensable step of any practical solution to entity alignment. 
However, nearly all prior studies have neglected dangling entities and assume there must be one-to-one entity mapping from the source KG to the target one \cite{sun2020benchmarking}. 
This assumption prevents prior methods from practically supporting the alignment between KGs in real-world scenarios. 
To fill the gap, \citet{sun_no_match} formally define a more practical problem setting
where a model needs to both determine whether each given source entity is a matchable one, as well as retrieve counterparts for the predicted matchable entities.



Although some preliminary attempts have been made to implement dangling-aware entity alignment \cite{sun_no_match}, the attempted methods still suffer from a major drawback, i.e., they only consider the first-order proximity (namely, pairwise cosine similarity) between source and target entities. 
However, to effectively discover dangling entities as outliers in the embedding representation, 
we argue that \textit{high-order proximity measures} should also be involved.
Fig.~\ref{fig:local_high_order} shows the intuition of using the high-order proximity for dangling entity detection.
Despite a fairly high cosine similarity, the source entity 4 is not the nearest neighbor of target entity B, indicating that 4 is likely to be dangling.
In contrast, 1 and A are mutual nearest neighbors even with a relatively low similarity, indicating that 1 is more likely to be matchable. 
Hence, in addition to the first-order proximity from source to target, the \textit{local high-order proximity} (e.g., the  second-order proximity\footnote{The second-order proximity of a  source entity $s$ is defined as aggregated cosine similarities between the nearest targets of $s$ and the nearest sources of these nearest targets.}
in the nearest neighbor subgraph) is also useful for detecting dangling entities.
In alignment learning, the previous works neglect global information since they merely optimize the entity-level alignment loss on labeled alignments without considering entity embedding distributions as shown in Fig.~\ref{fig:global_dist_a}.
In Fig.~\ref{fig:global_dist_b}, we show that a desirable dangling-aware model should align the \textit{global} distributions of matchable source and target entities (i.e., \textit{global high-order proximity} in an embedding distribution space), 
such that dangling entities in both KGs could appear as dissimilar outliers in both distributions.

Motivated by the above intuition, 
we propose a dangling-aware entity alignment framework based on \emph{\underline{m}ixed \underline{h}igh-order \underline{p}roximities} (\modelname).
\modelname considers both local and global high-order proximities to foster both dangling entity detection and matchable entity alignment.
We introduce the optimization of global high-order proximity measure as finding the Optimal Transport between 
matchable source entities and target entities. 
Through this optimization process, to facilitate dangling detection, \modelname also encourages a large distance between the dangling entity distribution and matchable entity distribution.
Additionally, to leverage the local high-order proximity, we propose a  dangling entity classifier which takes into account the second-order proximity in the nearest neighbor subgraph.
Furthermore, with the similar principle of local high-order proximity, we adopt an NCA (Neighborhood Component Analysis) loss \citep{goldberger2004neighbourhood,liu2021visual} for alignment learning to mitigate the hubness problem\footnote{The hubness problem is where some target entities dominantly appear as the nearest neighbors of many source entities.} \citep{radovanovic2010hubs},
which is severe in dangling-aware entity alignment as observed in our experiments.

Our main technical contributions to the studied problem are two-fold. 
First, the local high-order proximity (i.e., the second-order proximity in the nearest neighbor subgraph) is modeled to facilitate both dangling detection and alignment learning.
Second, we design 
the use of the global high-order proximity to align the distributions of matchable entities, 
therefore precisely separating the representations of dangling entities and matchable ones.
In addition, the techniques are model-agnostic and can be incorporated with various alignment methods
(e.g., MTransE \citep{chen_MTransE} or AliNet \citep{sun-etal-2020-knowledge}) and dangling detection methods (e.g., the marginal or background ranking \cite{sun_no_match}). 
Extensive experiments on DBP2.0 demonstrate its effectiveness and adaptiveness.

\section{Preliminary}
In this section, we provide the problem definition of dangling-aware entity alignment and briefly introduce previous methods for this problem.  

\subsection{Problem definition}
A KG is defined as $\mathcal{G} = (\mathcal{E}, \mathcal{R}, \mathcal{T})$, where $\mathcal{E}$ denotes a set of entities; $\mathcal{R}$ denotes a set of relations, and $\mathcal{T}\subset \mathcal{E}\times \mathcal{R}\times \mathcal{E}$ is a set of triples. 
Following the convention \cite{chen_MTransE}, we consider entity alignment between a source KG $\mathcal{G}_s = (\mathcal{E}_s, \mathcal{R}_s, \mathcal{T}_s)$ and a target KG $\mathcal{G}_t = (\mathcal{E}_t, \mathcal{R}_t, \mathcal{T}_t)$. 
Our study focuses on a more practical and challenging setting with dangling entities \citep{sun_no_match}.
In this setting, the training data contain a set of seed entity alignment $\mathcal{A} = \left\{\left(x_{s}, x_{t}\right) \in \mathcal{E}_{s} \times \mathcal{E}_{t} \| x_{s} \equiv x_{t}\right\}$ and a set of source dangling entities $\mathcal{D} \subset \mathcal{E}_{s}$ that has no counterparts in target KG.
After training, the model is required to first identify dangling entities and then find alignment for predicted matchable entities. 
This definition breaks the one-to-one assumption used in previous studies on the conventional setting \cite{sun2020benchmarking} and causes their methods to not be directly usable in our setting. 

\subsection{Dangling-aware entity alignment}
To the best of our knowledge, there is only one previous work \cite{sun_no_match} which has been attempted for dangling-aware entity alignment along with the proposing of this important problem.
This work also incorporates an embedding-based entity alignment technique (i.e., MTransE \citep{chen_MTransE} and AliNet \citep{sun-etal-2020-knowledge}) as the backbone, which 
learns alignment of KG embeddings based on the seed entity pairs.
Taking MTransE as an example, for each pair $(x_s, x_t) \in \mathcal{A}$, MTransE uses the learned embedding $\mathbf{x}_s$ and $\mathbf{x}_t$ to optimize a linear transformation matrix $\mathbf{M}$ by minimizing $||\mathbf{M}\mathbf{x}_s - \mathbf{x}_t||$. 
To detect dangling entities in the embedding space,
a margin ranking (MR) loss and a background ranking (BR) loss are experimented with,
both encouraging dangling entities to be isolated from others in the embedding space. 
MR sets a distance margin $\lambda$ to separate the dangling entity $x$ and its nearest neighbors by minimizing $\max \left(0, \lambda-\left\|\mathbf{M} \mathbf{x}-\mathbf{x}_{\mathrm{nn}}\right\|\right)$.
In like manner, BR treats dangling entities as the background of embedding space and learns to separate dangling entities and randomly-sampled other entities.

\section{Methodology}
\label{sec: method}
In this section, we introduce the techniques in our framework which
captures both local and global high-order proximities to collaboratively tackle dangling detection and entity alignment. 


\subsection{Global high-order proximity}
To leverage the global high-order proximity in an embedding space, in \modelname, we introduce a method based on Optimal Transport (OT) for globally aligning the distributions of matchable source and target entities. In addition, to facilitate dangling entity detection, the OT model encourages a large distribution distance 
between source- and target-KG dangling entities.
Intuitively, this strategy treats dangling entities as dissimilar parts of two embedding distributions, therefore tending to put dangling entities as outliers in the embedding space. 

\stitle{Optimal transport.} 
Let $\mathbf{s}$ and $\mathbf{t}$ be the distribution of transformed source-space embeddings $\mathbf{M}\mathbf{x}_s$ and target space embeddings $\mathbf{x}_t$, respectively.\footnote{Without loss of generality, we use a matrix M to transform embeddings from source KG to target KG.}
Intuitively, $\mathbf{s}$ should be similar with $\mathbf{t}$ 
if they represent matchable entities.
Meanwhile, to make dangling entities distinguishable, the distribution of transformed dangling entity vectors $\mathbf{M}\mathbf{x}_d$ should be different from $\mathbf{t}$. 
The discrepancy between $\mathbf{s}$ and $\mathbf{t}$ can be represented as a Wasserstein distance  which is one type of OT distance \citep{peyre2019computational_OT}: 
\begin{align}
\label{eq: ot_1}
    \mathcal{D}_{c}(\mathbf{s}, \mathbf{t})=\inf _{\gamma \in \Pi(\mathbf{s}, \mathbf{t})} \mathbb{E}_{(\mathbf{x}, \mathbf{y}) \sim \boldsymbol{\gamma}}[c(\mathbf{x}, \mathbf{y})],
\end{align}
where $\Pi(\mathbf{s}, \mathbf{t})$ is the set of all possible joint distributions $\gamma(\mathbf{s}, \mathbf{t})$ and $c(\mathbf{x}, \mathbf{y})$ denotes the cost function describing the distance between $\mathbf{x}$ and $\mathbf{y}$. 
Then the Wasserstein distance $\mathcal{D}_{c}(\mathbf{s}, \mathbf{t})$ denotes the cost of the optimal transport plan. 

However, the infimum to calculate  $\mathcal{D}_{c}(\mathbf{s}, \mathbf{t})$ is highly intractable \citep{arjovsky2017wasserstein}. 
To handle this, the Kantorovich-Rubinstein duality points out that Eq. \eqref{eq: ot_1} can be transformed to: 
\begin{equation}
    \mathcal{L}_{o t}(\mathbf{s}, \mathbf{t})=\frac{1}{K} \sup _{\|f\|_{L} \leq K} \mathbb{E}_{x \sim \mathbf{t}}[f(x)]-\mathbb{E}_{x \sim \mathbf{s}}[f(x)], 
\end{equation}
where the supremum is over all possible K-Lipschitz functions $f$.
As \citet{arjovsky2017wasserstein} point out that optimizing Wasserstein GAN (WGAN) can be used to solve this optimal transport problem, we utilize WGAN in our study. 
Specifically, we adopt a MLP to approximate the function $f$ (called as critic $D$) since neural networks are universal approximators \citep{hornik1989multilayer}. 
The objective of the critic is defined as follows: 
\begin{equation}
    \max_{D}  \mathbb{E}_{y \sim \mathbf{t}}\left[f_{D}(y)\right]-\mathbb{E}_{x \sim \mathbf{s}}\left[f_{D}\left(\mathbf{M} x\right)\right]. 
\end{equation}
Thus, the critic $D$ aims to distinguish transformed source embeddings from target embeddings. 
In contrast, the transformation matrix $\mathbf{M}$ tries to minimize the distance between the two sets of embeddings. The objective to optimize $\mathbf{M}$ is defined as: 
\begin{align}
    & \min _{\mathbf{M}} \mathbb{E}_{y \sim \mathbf{t}}\left[f_{D}(y)\right] -\mathbb{E}_{x \sim \mathbf{s}}\left[f_{D}\left(\mathbf{M} x\right)\right] \nonumber \\
    & = \min _{\mathbf{M}} -\mathbb{E}_{x \sim \mathbf{s}}\left[f_{D}\left(\mathbf{M} x\right)\right].
\end{align}
Therefore, conducting entity alignment with the consideration of whole embedding distributions is converted to the problem of optimizing a WGAN. 

So far, only the distribution of matchable source entity embeddings and that of target entity embeddings are considered, whereas the distribution of dangling entity embeddings is neglected. 
Therefore, to tailor the optimization problem for dangling entities, we adopt an additional objective for the transformation matrix $\mathbf{M}$: 
\begin{align}
    & \max _{\mathbf{M}} \mathbb{E}_{y \sim \mathbf{t}}\left[f_{D}(y)\right] -\mathbb{E}_{x \sim \mathbf{d}}\left[f_{D}\left(\mathbf{M} x\right)\right] \nonumber \\
    & = \min _{\mathbf{M}} \mathbb{E}_{x \sim \mathbf{d}}\left[f_{D}\left(\mathbf{M} x\right)\right], 
\end{align}
where $\mathbf{d}$ denotes the distribution of dangling entity embeddings. Hence, the transformation matrix $\mathbf{M}$ is enforced to maximize the difference between the distribution of transformed dangling embeddings and that of target entity embeddings, which can make dangling entities more distinguishable. 


\subsection{Local high-order proximity}
\begin{figure}
    \centering
    \begin{subfigure}[b]{0.45\textwidth}
    \centering
    \includegraphics[width=\textwidth]{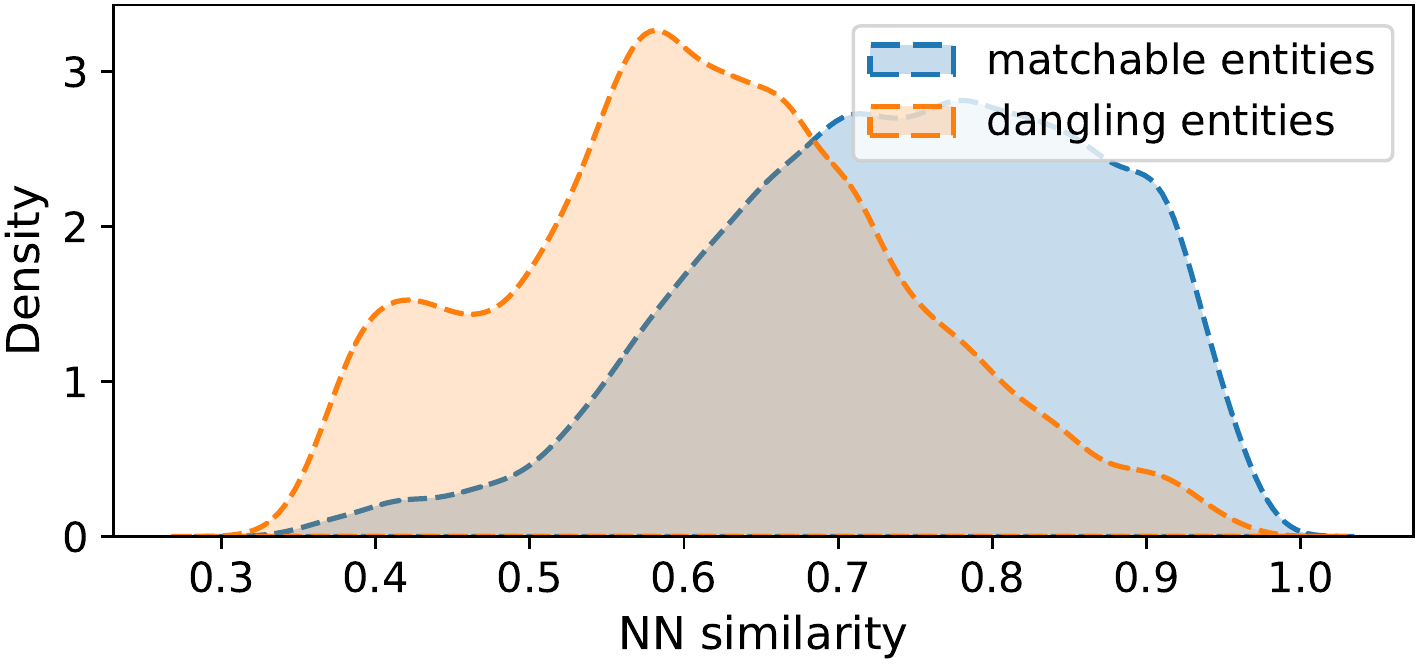}
    \caption{Proximity distribution of nearest entities.}
    \label{fig:sim_dist_1st}
    \end{subfigure}
    \begin{subfigure}[b]{0.45\textwidth}
    \centering
    \includegraphics[width=\textwidth]{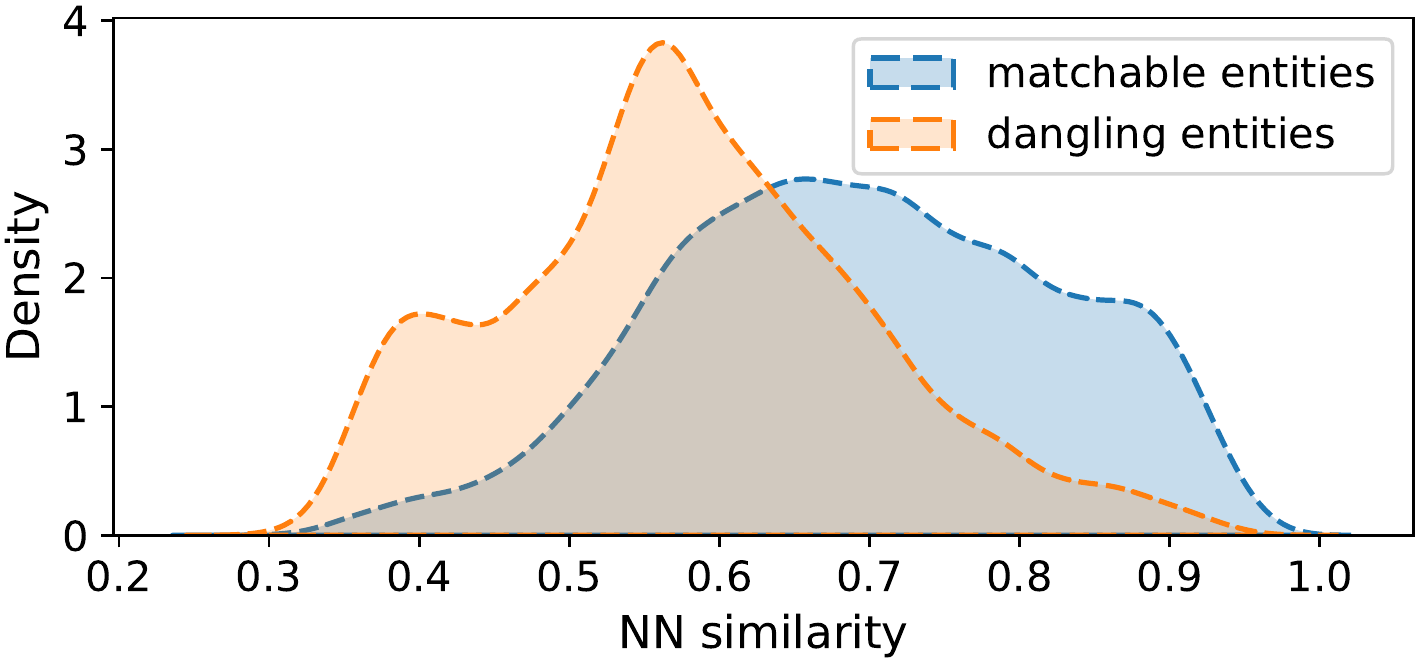}
    \caption{Proximity distribution of the second nearest entities.}
    \label{fig:sim_dist_2nd}
    \end{subfigure}
    \caption{Second-order proximity distributions.}
\end{figure}
In addition to the global proximity measure, \modelname also captures local high-order proximity measures in the nearest neighbor subgraph.
In contrast, the previous work~\citep{sun_no_match} merely uses the first-order proximity between an individual source entity $s$ and its nearest target entity to decide whether $s$ should be dangling.
However, apart from the first-order proximity, the second-order proximity measure could be informative as well for detecting dangling entities. 

Furthermore, we verify the above hypothesis empirically using the previous work.
From the nearest target entity $t$ of a given source entity, we obtain the cosine similarities between $t$ and its top 2 nearest source entities as the second-order proximity measures, and plot the proximity distributions in Fig. \ref{fig:sim_dist_1st} and \ref{fig:sim_dist_2nd} for the first and second nearest entities, respectively. 
Fig. \ref{fig:sim_dist_1st} shows that the second-order proximity 
between matchable entities and their nearest neighbors appear as a very different distribution in comparison to that of the proximity between dangling entities and their nearest neighbors.
Fig. \ref{fig:sim_dist_2nd} demonstrates a similar 
observation for the proximity distributions of the second nearest entities. 
Therefore, the second-order proximities are informative and should be used for distinguishing dangling entities, meanwhile combining proximity measures that consider multiple neighboring entities is more useful than a single similarity measure on only the nearest entity.

To this end, we design a dangling entity classifier using both first-order and second-order proximity measures as the input. 
From the perspective of a given source entity $s$, we conduct nearest neighbor search to obtain top $k$ nearest target entities $\{t_1, ..., t_k\}$ and their proximity score vector $\mathbf{d}_1 = [d_{st_1}, ..., d_{st_k}] \in \mathbb{R}^{1 \times k}$. 
The proximity $d_{st}$ is measured by the cosine similarity between transformed source embedding $\mathbf{Mx}_s$ and target entity embedding $\mathbf{x}_{t}$: 
\begin{equation}
    d_{st_1} = \Big\langle\frac{\mathbf{Mx}_s}{\|\mathbf{Mx}_s\|_{2}}, \frac{\mathbf{x}_{t}}{\|\mathbf{x}_{t}\|_{2}}\Big\rangle 
\end{equation}
After getting the first-order proximity vector $\mathbf{d}_1$ between the source and target, through the reverse direction (target KG to source KG), we can further obtain the second-order proximity vector.
Specifically, we retrieve top $m$ nearest source entities $\{s_1, ..., s_m\}$ of each target entity $t$ in $\{t_1, ..., t_k\}$.
Accordingly, through the target entity $t$, the second-order proximity measures with regard to the $m$ retrieved source entities 
are obtained as the vector $\mathbf{d}_{t} = [d_{s_1t}, ..., d_{s_mt}] \in \mathbb{R}^{1\times m}$. 
Subsequently, we can collect $\{\mathbf{d}_{t_1} ... \mathbf{d}_{t_k}\}$ and concatenate them as the whole second-order proximity vector $\mathbf{d}_2 = \left[\mathbf{d}_{t_1} || ... ||  \mathbf{d}_{t_k}\right] \in \mathbb{R}^{1\times km}$. 

To utilize both second-order and first-order information, the whole proximity distribution vector is constructed as $\mathbf{d} = [\mathbf{d_1} || \mathbf{d_2}] \in \mathbb{R}^{1 \times (k+1)m}$. 
In this way, we 
use the distribution as profile of the neighborhood of a source entity $s$, then we adopt a simple feed-forward neural network (FNN) binary classifier to determine whether $s$ is dangling. 
The probability of $s$ being a dangling entity can be calculated as $p(y=1 | s) = \text{sigmoid}(\text{FNN}(\mathbf{d}))$. 
Define $\mathcal{D}$ and $\mathcal{A}$ to be the training set of dangling entities and that of matchable entities, respectively. We minimize the binary cross-entropy loss: 
\begin{equation}
\begin{aligned} 
\mathcal{L}_{s}= -&\frac{1}{\left|\mathcal{D} \cup \mathcal{A}\right|}\sum_{s \in \mathcal{D} \cup \mathcal{A}}\left(y_{s} \log (p(y=1|s))\right. \\ 
            &\left.+\left(1-y_{s}\right) \log (1-p(y=1|s))\right) 
\end{aligned}
\end{equation}

\stitle{NCA loss.}
With the similar principle of local high-order proximity, \modelname adopts an additional Neighbor Component Analysis (NCA) loss \citep{liu2021visual} to mitigate the hubness problem.
The hubness problem can be more severe in dangling-aware entity alignment as dangling entities might be aligned to some certain hubs if they are not detected as dangling.
The NCA loss measures importance of samples and punishes hard negative pairs based on the proximities. 
Given the set of seed entity alignments $\{(x_s,x_t) \in \mathcal{E}_s \times \mathcal{E}_t\}$, let $\mathbf{S}$ be the cosine similarity matrix between source and target entity embeddings $\mathbf{E}_1$ and $\mathbf{E}_2$. The NCA loss can be defined as follows: 
\begin{equation}
\begin{aligned}
    \mathcal{L}_{NCA} = \frac{1}{N}\sum_{i=1}^{N}\Big(\frac{1}{\alpha}\log (1 + \sum_{m \neq i} e^{\alpha\mathbf{S}_{im}}) + \Big. \\
    \Big. \frac{1}{\alpha} \log(1 + \sum_{n\neq i}e^{\alpha\mathbf{S}_{ni}}) - \log (1 + \beta e^{\mathbf{S}_{ii}}) \Big),
\end{aligned}
\end{equation}
where $\mathbf{S}_{ii}$ denotes the proximity of the i-th positive pair (i.e., the i-th source entity and the i-th target entity); $\alpha, \beta$ are temperature hyper-parameters; and $N$ is the number of positive pairs in the mini-batch. 


\subsection{Learning and inference}
Note that our techniques are used to improve existing first-order methods.
\modelname optimizes the entity alignment component and the dangling detection component alternately. 
For entity alignment, besides an entity-level loss (e.g., MTransE), we 
first train WGAN for optimal transport and then optimize the NCA loss $\mathcal{L}_{NCA}$. 
For dangling detection, besides a first-order objective (e.g., a marginal ranking loss) used in \citet{sun_no_match}, we train our dangling classifier for detection. 
In the inference phase, for each source entity, the dangling entity classifier provides a probability score and uses a probability threshold to decide whether an entity is dangling, where the threshold is set as the average probability.
After this dangling detection process, the predicted dangling entities are excluded from being aligned.
Then, in the alignment process, \modelname conducts nearest neighbor search to find the alignment in the target KG embedding space for each of the rest matchable source entities. 

\section{Experiments}
In this section, we report our experiments to show the effectiveness of \modelname.
We describe the evaluation settings in Sec.~\ref{sec: experimental_settings}, 
and present the results in two alignment settings separately in Sec.~\ref{sec: consolidated evaluation} and~\ref{sec: relaxed evaluation}.
We conduct an ablation study and demonstrate that \modelname can mitigate the hubness problem in Sec.~\ref{sec: ablation}, followed by a case study to show the importance of local high-order proximity in Sec.~\ref{sec: case_study}. 

\subsection{Experimental settings}
\label{sec: experimental_settings}
We use two evaluation settings as suggested by \citet{sun_no_match}. 
The first one is \textit{consolidated evaluation} which 
requires a model to first detect and remove dangling entities, and then conduct alignment search for the rest of entities. 
The performance of dangling entity detection is also evaluated in this setting. 
Besides, 
a simplified \textit{relaxed evaluation setting} 
seeks to test the performance of alignment alone without involving dangling source entities in the test set.
In this setting, the effect of dangling detection on entity alignment can be evaluated.

\stitle{Evaluation protocol.} For the \textit{relaxed setting}, the counterpart list is selected by the Nearest Neighbor (NN) search in the embedding space for each source entity. To assess the ranking list, we use mean reciprocal rank (MRR), Hits@$1$ and Hits@$10$ (hereinafter H@$1$ and  H@$10$) as metrics. 
Higher values indicate better performance. 

For the \textit{consolidated setting}, we evaluate the performance of both dangling entity detection and entity alignment
using precision, recall, and F1 score, 
following \citet{sun_no_match}.\footnote{Note that H@$1$, H@$10$ and MRR are not applicable to this entity alignment in the consolidated setting.}  
In this setting,
only the source entities that are correctly predicted as matchable are sent to the NN search and the nearest counterpart is evaluated.
Particularly, incorrect dangling detection (i.e., a matchable entity is wrongly predicated as dangling or a dangling entity is predicted as matchable) will propagate an error case to the alignment process.
We refer to this practical entity alignment as \textit{two-step entity alignment}. 

\stitle{Dataset.}
We use the cross-lingual dangling-aware entity alignment dataset DBP2.0 \cite{sun_no_match},
which is constructed using multilingual DBpedia \citep{lehmann2015dbpedia}.
There are three language pairs (ZH-EN, JA-EN, FR-EN) in DBP2.0 and two alignment directions are considered for each pair. 
We follow its data splits where 30\% dangling entities are for training, 20\% for validation, and 50\% for test.
The dataset statistics are reported in Appx.~\ref{sec:appendix}.

\begin{table*}[!t]
	\centering
	\resizebox{.999\textwidth}{!}
	{\small
	\setlength{\tabcolsep}{3pt}
		\begin{tabular}{lcccccccccccccccccc}
			\toprule
            Methods &
			\multicolumn{3}{c}{ZH-EN} & \multicolumn{3}{c}{EN-ZH} & \multicolumn{3}{c}{JA-EN} & \multicolumn{3}{c}{EN-JA} &  \multicolumn{3}{c}{FR-EN} & \multicolumn{3}{c}{EN-FR}\\
			\cmidrule(lr){2-4} \cmidrule(lr){5-7} \cmidrule(lr){8-10} \cmidrule(lr){11-13} \cmidrule(lr){14-16} \cmidrule(lr){17-19}
			& Prec. & Rec. & F1 & Prec. & Rec. & F1 & Prec. & Rec. & F1 & Prec. & Rec. & F1 & Prec. & Rec. & F1 & Prec. & Rec. & F1 \\ 
			\midrule
			MR & .781 & .702 & .740 & .866 & .675 & .759 & .799 & .708 & .751 & .864 & .653 & .744 & .482 & .575 & .524 & .639 & .613 & .625 \\
			BR & \textbf{.811} & .728 & .767 & \textbf{.892} & .700 & .785 & \textbf{.816} & .733 & .772 & \textbf{.888} & .731 & .801 & .539 & .686 & .604 & .692 & .735 & .713 \\
            \midrule
             \modelname + MR& .784 & \textbf{.831} & \textbf{.807} & .858 & \textbf{.861} & \textbf{.859} & .815 & \textbf{.791} & \textbf{.803} & .865 & \textbf{.852} & \textbf{.858} & \textbf{.580} & \textbf{.724} & \textbf{.644} & \textbf{.707} & \textbf{.749} & \textbf{.727} \\
             \modelname + BR & .758 & .815 & .785 & .832 & .847 & .839 & .783 & .785 & .784 & .834 & .848 & .841 &  .569 & .706 & .635 & .685 & .747 & .714\\
			\bottomrule
	\end{tabular}}
	\caption{Dangling entity detection results on DBP2.0. MR refers to marginal ranking and BR refers to the background ranking. The base alignment model is MTransE. More results based on AliNet are in Appx. Tab.~\ref{tab:detection_alinet}.}
	\label{tab:detection}
\end{table*}

\begin{table*}[!t]
	\centering
	\resizebox{.999\textwidth}{!}
	{\small
	\setlength{\tabcolsep}{3pt}
		\begin{tabular}{lcccccccccccccccccc}
			\toprule
			Methods &
			\multicolumn{3}{c}{ZH-EN} & \multicolumn{3}{c}{EN-ZH} & \multicolumn{3}{c}{JA-EN} & \multicolumn{3}{c}{EN-JA} &  \multicolumn{3}{c}{FR-EN} & \multicolumn{3}{c}{EN-FR}\\
			\cmidrule(lr){2-4} \cmidrule(lr){5-7} \cmidrule(lr){8-10} \cmidrule(lr){11-13} \cmidrule(lr){14-16} \cmidrule(lr){17-19}
			& Prec. & Rec. & F1 & Prec. & Rec. & F1 & Prec. & Rec. & F1 & Prec. & Rec. & F1 & Prec. & Rec. & F1 & Prec. & Rec. & F1 \\ 
			\midrule
			MR & .302 & .349 & .324 & .231 & .362 & .282 & .313 & .367 & .338 & .227 & .366 & .280 & .260 & .220 & .238 & .213 & .224 & .218 \\
			BR & .312 & .362 & .335 & .241 & \textbf{.376} & .294 & .314 & .363 & .336 & .251 & .358 & .295 & .265 & .208 & .233 & .231 & .213 & .222 \\
            \midrule
            \modelname + MR & \textbf{.400} & \textbf{.363} & \textbf{.381} & \textbf{.375} & .372 & \textbf{.373} & \textbf{.378} & \textbf{.394} & \textbf{.386} & \textbf{.371} & \textbf{.384} & \textbf{.377} & \textbf{.310} & \textbf{.249} & \textbf{.276} & .266 & \textbf{.260} & \textbf{.263} \\
            \modelname + BR & .393 & .347 & .368 & .347 & .331 & .339 & .374 & .372 & .373 & .359 & .344 & .352 & .290 & .235 & .259 & \textbf{.269} & .239 & .253\\
			\bottomrule
	\end{tabular}}
	\caption{Two-step entity alignment results on DBP2.0. The base alignment model is MTransE.}
	\label{tab:ent_alignment}
\end{table*}

\stitle{Baselines.}
To the best of our knowledge, the framework with a dangling detection module proposed in \citet{sun_no_match} is the only study on dangling-aware entity alignment. 
It includes three dangling detection techniques: (i) NN classification, (ii) marginal ranking (MR), and (iii) background ranking (BR). 
As the NN classification performs much worse than others, we choose MR and BR as baselines.
For a fair comparison with \cite{sun_no_match}, we use the same 
base alignment model MTransE \cite{chen_MTransE}. 
The results using AliNet \cite{AliNet} \footnote{AliNet performs worse than MTransE on dangling-aware entity alignment as found by \citet{sun_no_match}.} as a base 
are presented in Appx.~\ref{appendix: more_experiments}. 
Note that our methods are model-agnostic and can be incorporated with any detection and alignment methods.
The entity alignment models that consider side information are left for future work.

\stitle{Model configuration.} 
In \modelname, aside from our proposed components as described in Sec.~\ref{sec: method}, we have a base alignment module (e.g., MTransE) and a base dangling entity loss (e.g., MR) as in \citet{sun_no_match}. 
For KG embeddings and model weights, we use Xavier initialization \citep{glorot2010understanding} and optimize them using Adam optimizer \citep{kingma2014adam}.
The number of hidden units in the dangling entity classifier is 128. 
The number of nearest targets $k$ and nearest sources $m$ are set as 5. 
The learning rate is set to 0.001 for all components except WGAN where the learning rate is 5e-5 for three objectives. 
To terminate training, early stopping is used based on the F1 score of two-step entity alignment on validation set. 
The computational environment and other configuration details are reported in Appx.~\ref{appx:computation} and \ref{appx:parameters}.

\subsection{Consolidated evaluation}
\label{sec: consolidated evaluation} 

\stitle{Dangling entity detection.} 
According to the results in Tab.~\ref{tab:detection}, no matter which base dangling detection loss we adopt, \modelname consistently achieves better F1 scores compared with the corresponding baseline by \citet{sun_no_match} without our proposed techniques.
In terms of the recall, \modelname also outperforms baselines with a large margin, which indicates that our framework has a better coverage to find more dangling entities. 
With better recalls, \modelname has the same level or slightly worse precision compared with baselines. 
But our higher recall and F1 scores in dangling detection imply that more predicted matchable source entities would enter two-step entity alignment, which can improve the final alignment performance.
Comparing \modelname + MR and \modelname + BR, we can see that the MR variant is generally better than the BR variant. 
This is because MR considers the similarity between a source and its nearest neighbor, which can benefit the learning of local high-order proximity in \modelname. 
In summary, \modelname demonstrates superior effectiveness for detecting dangling entities. 

\begin{table*}[!t]
	\centering
	\resizebox{.999\textwidth}{!}
	{\small
	\setlength{\tabcolsep}{2pt}
		\begin{tabular}{lcccccccccccccccccc}
			\toprule
			\multirow{2}{*}{Methods} &
			\multicolumn{3}{c}{ZH-EN} & \multicolumn{3}{c}{EN-ZH} & \multicolumn{3}{c}{JA-EN} & \multicolumn{3}{c}{EN-JA} & \multicolumn{3}{c}{FR-EN} & \multicolumn{3}{c}{EN-FR}\\
			\cmidrule(lr){2-4} \cmidrule(lr){5-7} \cmidrule(lr){8-10} \cmidrule(lr){11-13} \cmidrule(lr){14-16} \cmidrule(lr){17-19}
			& H@1 & H@10 & MRR & H@1 & H@10 & MRR & H@1 & H@10 & MRR & H@1 & H@10 & MRR & H@1 & H@10 & MRR & H@1 & H@10 & MRR \\ 
			\midrule
			MTransE & {.358} & {.675} & {.463} & .353 & {.670} & {.461} & {.348} & {.661} & {.453} & .342 & {.670} & .452 & {.245} & {.524} & {.338} & {.247} & {.531} & {.342} \\
			\;\;w/ MR & .378 & .693 & .487 & .383 & .699 & .491 & .373 & .686 & .476 & .374 & .707 & .485 & .259 & .541 & .348 & .265 & .553 & .360 \\
			\;\;w/ BR & .360 & .678 & .468 & .357 & .675 & .465 & .344 & .660 & .451 & .346 & .675 & .456 & .251 & .525 & .342 & .249 & .531 & .343 \\
			\midrule
            \modelname + MR & \textbf{.418} & \textbf{.727} & \textbf{.523} & \textbf{.404} & \textbf{.724} & \textbf{.513} & \textbf{.408} & \textbf{.730} & \textbf{.517} & \textbf{.410} & \textbf{.747} & \textbf{.524} & .274 & .568 & .371 & \textbf{.274} & .566 & \textbf{.370} \\
            \modelname + BR & .412 & .718 & .517 & .396 & .714 & .505 & .400 & .727 & .511 & .400 & .728 & .511 & \textbf{.278} & \textbf{.574} & \textbf{.376} & .272 & \textbf{.569} & \textbf{.370} \\
			\bottomrule
	\end{tabular}}
	\caption{Entity alignment results in the relaxed setting on DBP2.0.}
	\label{tab:synthetic_ent_alignment}
\end{table*}

\stitle{Two-step entity alignment.}
The results of two-step alignment are shown in Tab.~\ref{tab:ent_alignment}.
In general, \modelname again consistently offers better F1 scores than baseline methods.  
The relative improvement ranges from 11\% to 32\%.
The improvement can be partly attributed to the more accurate dangling entity detection performance, and thus less error is propagated to the alignment process. 
In contrast, baselines may try to align many dangling entities, which leads to lower performance on two-step alignment. 
As \modelname with MR outperforms \modelname + BR in dangling detection, \modelname + MR also achieves better performance in two-step alignment. 
This indicates that dangling entity detection is of importance on the dangling-aware entity alignment problem since it has strong effects on the performance of two-step alignment. 



\begin{table}[!t]
	\centering
	\resizebox{.9\columnwidth}{!}
	{
	\setlength{\tabcolsep}{3pt}
		\begin{tabular}{lcccc}
			\toprule
            \multirow{2}{*}{Methods} &
			\multicolumn{2}{c}{Dangling detection} & \multicolumn{2}{c}{Two-step alignment}\\
			\cmidrule(lr){2-3} \cmidrule(lr){4-5}
			 & F1 & $\Delta$ & F1 & $\Delta$  \\ 
			\midrule
			\modelname & .807 & 0 & .381 & 0 \\
			\midrule
		    \,\,- Dangling cls. & .752 & -.055 & .339  & -.042 \\
		    \,\,- OT &  .789 & -.018 & .369 & -.012 \\
		    \,\,- NCA & .803 & -.004 & .361 & -.020 \\
			\bottomrule
	\end{tabular}}
	\caption{Ablation study in the consolidated setting on ZH-EN. We remove each technique and report the performance decline $\Delta$ compared with the full \modelname.}
	\label{tab: ablation}
\end{table}

\subsection{Relaxed evaluation }
\label{sec: relaxed evaluation} 
Tab.~\ref{tab:synthetic_ent_alignment} shows the results of relaxed evaluation. 
This setting only considers matchable source entities in the test phase to investigate how our framework affects the alignment learning of these entities. 

Generally, \modelname offers better performance than baselines on all language pairs in terms of all metrics.
This indicates that dangling awareness captured by \modelname further helps with a more precise alignment.
The improvement can also be partly attributed to the alleviated hubness problem by the NCA loss which we investigate more in Sec.~\ref{sec: ablation}.
Comparing two variants of \modelname, we can see that \modelname + MR usually outperforms the BR variants on most language pairs except for FR-EN. 
The reason could be that FR-EN has more entities and
only with sufficient data BR can effectively separate dangling entities from randomly sampled target entities, while MR is not sensitive to data volume. 

\subsection{Ablation study}
\label{sec: ablation}

To investigate the effectiveness of each module in \modelname, we conduct an ablation study on the consolidated setting and show the results in Tab.~\ref{tab: ablation}. 
Compared with the full version \modelname, removing any component causes the degraded performance.  
Specifically, by removing the dangling classifier, the F1 score of dangling detection drops 0.055, which also leads to a large performance drop on two-step alignment. 
This indicates that the local high-order proximity is useful for dangling detection. 
Removing OT decreases the F1 scores on both detection and two-step alignment, showing the effectiveness of globally aligning distributions. 
Lastly, leaving the NCA loss out makes the F1 score of two-step alignment decrease 0.02 compared with \modelname, because using the NCA loss reduces the extent of hubness, as discussed below. 

\begin{figure}[!t]
    \centering
    \includegraphics[width=0.95\columnwidth]{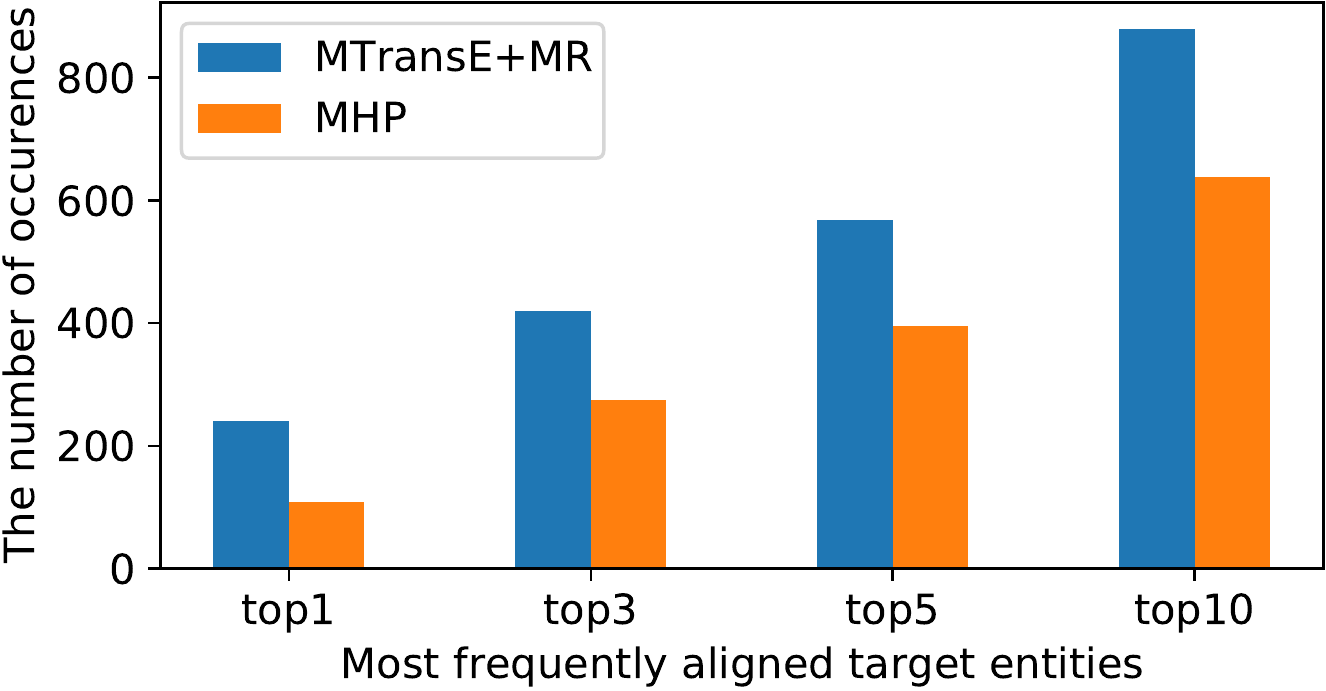}
    \caption{The number of total occurrences of most frequently aligned target entities on ZH-EN.}
    \label{fig:hubness}
\end{figure}
\label{sec: similarity distribution} 

\stitle{Hubness problem.}
To examine whether the NCA loss reduces the hubness problem, we 
list a set of most frequently aligned (target) entities, 
and observe how frequently they appear as the nearest neighbor of other entities in the embedding space.
We compare \modelname with the MTransE + MR variant used by \citet{sun_no_match}. 
As shown in Fig.~\ref{fig:hubness}, the most frequently aligned target entity (i.e., top 1) appears over 200 times as the nearest neighbor using the baseline, whereas it only appears around 100 times using \modelname. 
A similar phenomenon is also observed for the top 3, top 5, and top 10 frequently aligned target entities. 
This indicates that the hubness problem is mitigated by using NCA.

\subsection{Case study}
\label{sec: case_study}
\begin{figure}
    \centering
    \includegraphics[width=0.9\columnwidth]{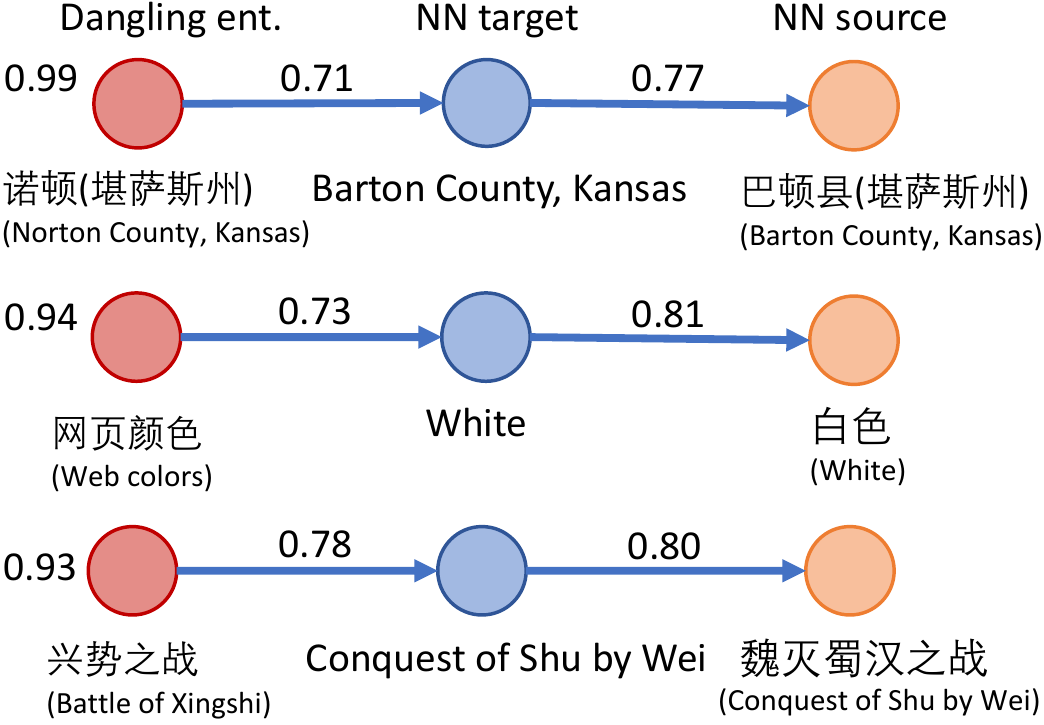}
    \caption{Case study on ZH-EN where some dangling entities wrongly predicted as matchable by the previous first-order method can be correctly predicted as dangling with high probabilities via \modelname. Arrows point from an entity to its NN in the other KG. The scores above arrows denote cosine similarities and those beside dangling ent. are probabilities of dangling by \modelname.}
    \label{fig:case_study}
\end{figure}

To further investigate the superiority of \modelname, we provide a case study on ZH-EN comparing \modelname with the previous method. 
Fig.~\ref{fig:case_study} shows that, the previous method predicts some dangling source entities as matchable based on their 
high cosine similarities (i.e., > 0.7) to their nearest target entities. 
Each dangling entity and its corresponding nearest target entity are different but share similar meanings (e.g., are both war events in ancient China or locations).
However, the nearest targets prefer other source entities with higher similarities.
Using this second-order proximity information, \modelname correctly detects these dangling entities with high probability scores (i.e., > 0.9). 

Additionally, Tab.~\ref{tab: appendix_case_study} demonstrates more dangling entities which are not correctly detected by the previous method~\citep{sun_no_match}. 
Most of the dangling entities are aligned to some similar counterparts sharing the same attribute. 
For example, the dangling entity and its nearest target entity are both colleges, theoretical physicists, or political parties. However, from the view of the nearest target entity, it prefers other nearest neighbors on source KG. 
Such second-order proximity information cannot be captured by the previous method, which causes those dangling entities not able to be detected. 
In contrast, \modelname can successfully detect those dangling entities with high probabilities.
This shows the effectiveness of \modelname and the informativeness of the second-order proximity. 

\begin{CJK*}{UTF8}{gbsn}
\begin{table*}[!t]
    \centering\resizebox{.999\textwidth}{!}{\setlength{\tabcolsep}{8pt}{\Large
    \begin{tabular}{cccccc}
        \toprule
         Dangling entity & Cls. Prob.& The nearest target & Cosine Sim. & The nearest source & Cosine Sim. \\
         \midrule
        哥伦比亚国际学院 \small{(Columbia International College)} &	0.95 &	University of Ottawa & 0.67 & 渥太华大学 \small{(University of Ottawa)} & 	0.80 \\ 
         丁肇中 \small{(Samuel C. C. Ting)} & 0.91 & George Uhlenbeck & 0.65 & 乔治·乌伦贝克 \small{(George Uhlenbeck)} & 0.78 \\
         美国国会地铁 \small{(Congressional Subway)} & 0.99 & United States Congress & 0.67 & 美国国会 \small{(United States Congress)} & 0.75\\ 
         王豫元 \small{(Larry Wang)} & 1.00 &	 Wu Den-yih &  0.65 &	 吴敦义 \small{(Wu Den-yih)} & 0.91 \\
         新生党 \small{(Japan Renewal Party)} & 1.00 &  Democratic Party of Japan & 0.64 & 	 民主党 (日本) \small{(Democratic Party of Japan)} & 0.85 \\ 
         意大利裔澳洲人 \small{(Italian Australians)} & 0.99 & 	 Chinese Australians 	 & 0.64 & 澳大利亚华人 \small{(Chinese Australians)} & 0.71 \\ 
         新加坡发展部 \small{(Ministry of Development (Singapore))} &	0.93 & Ministry of Transport (Singapore) & 0.72 & 林瑞生 \small{(Lim Swee Say)} & 0.75 \\ 
         \bottomrule
    \end{tabular}}}
    \caption{Some dangling source entities wrongly predicted as matchable by the previous method, while \modelname predicts them as dangling with high probabilities. Cls. Prob. denotes the probabilities of dangling generated by \modelname. The fourth column denotes the cosine similarity between the dangling entity and its nearest target. The nearest source is the nearest neighbor of the nearest target on the source KG. The last column denotes the cosine cosine similarity between the nearest target and its nearest source.}
    \label{tab: appendix_case_study}
\end{table*}
\end{CJK*}


\section{Related Work}

\vspace{-2mm}
\stitle{Entity alignment.}
Embedding-based entity alignment methods seek to find identical entities between KGs in their embedding spaces.
Such a method encodes each KG into an embedding space
and capture entity alignment by learning a linear mapping between embedding spaces \cite{chen_MTransE} or directly infer the embedding proximity in a shared space \cite{JAPE}.
Existing studies mainly fall into two lines of improving the embedding representations.
The first line exploits better graph encoders to improve embedding learning \cite{sun2018bootstrapping,wang-etal-2018-cross-lingual,cao-etal-2019-multi,sun-etal-2020-knowledge,AliNet,Fey2020Deep}.
The second group considers the side information of entities \cite{chen2018co,AttrE,MultiKE,GMNN,WangZhiChun20,BERTINT,RDGCN,HMAN,AttrGNN,liu2021visual}.
Interested readers can refer to the recent surveys \cite{sun2020benchmarking, zeng2021comprehensive}.
Note that prior methods nearly all assume one-to-one perfect match exists between two KGs, without considering dangling entities.

Recently, \citet{sun_no_match} have proposed a new problem setting, i.e., danging-aware entity alignment,
which is more practical as dangling entities naturally exist in real-world KGs.
This problem setting requests a model to both detect dangling entities and align matchable ones.
As the pioneering work, \citet{sun_no_match} propose three baseline methods (i.e., marginal ranking, background ranking, and nearest neighbor classification) based on the nearest neighbor of source entities.
Thus, these methods only rely on the first-order proximity, which is the major difference with \modelname. 


\stitle{Optimal transport.}
Optimal transport (OT) aims to find the plan with minimal transportation cost for changing one distribution to another distribution, which naturally provides a way to align two distributions. 
\citet{arjovsky2017wasserstein} use the Wasserstein distances to recast the learning of generative adversarial network (GAN) as a transportation problem.
OT has been widely used in other applications like text generation \citep{chen2018adversarial} and graph matching \citep{xu2019gromov}. 
\citet{pei2019improving_OT} formalize entity alignment as OT in the conventional setting,
which however only considers one-to-one alignment between matchable entities.
We instead leverage OT to identify dissimilar parts of embedding distributions to detect dangling entities, meanwhile using OT only as one of the three high-order measures for alignment.

\section{Conclusion}
In this paper, we propose a framework, \modelname, with mixed high-order proximities for dangling-aware entity alignment. 
\modelname captures the local high-order proximity via a dangling classifier based on both the first- and second-order proximities. 
Additionally, we propose a Optimal Transport based method considering the global high-order proximity to facilitate both dangling detection and entity alignment. 
Comprehensive experiments on two alignment settings show the effectiveness of utilizing mixed high-order proximities.
Furthermore, our extensive ablation study demonstrates the effectiveness of each technique.

\section*{Acknowledgments}
We appreciate the anonymous reviewers for their insightful comments and suggestions. 
Juncheng Liu and Xiaokui Xiao are supported by the Ministry of Education, Singapore (Grant No. MOE2018-T2-2-091). 
Muhao Chen is supported by the National Science Foundation of United States Grant IIS 2105329, and by the DARPA MCS program under Contract No. N660011924033 with the United States Office Of Naval Research.
The views and conclusions contained in this paper are those of the authors and should not be interpreted as representing any funding agencies.
\bibliography{anthology,custom}

\begin{thebibliography}{36}
\expandafter\ifx\csname natexlab\endcsname\relax\def\natexlab#1{#1}\fi

\bibitem[{Arjovsky et~al.(2017)Arjovsky, Chintala, and
  Bottou}]{arjovsky2017wasserstein}
Mart{\'{\i}}n Arjovsky, Soumith Chintala, and L{\'{e}}on Bottou. 2017.
\newblock \href {http://proceedings.mlr.press/v70/arjovsky17a.html}
  {Wasserstein generative adversarial networks}.
\newblock In \emph{Proceedings of the 34th International Conference on Machine
  Learning (ICML)}, pages 214--223.

\bibitem[{Cao et~al.(2019)Cao, Liu, Li, Liu, Li, and
  Chua}]{cao-etal-2019-multi}
Yixin Cao, Zhiyuan Liu, Chengjiang Li, Zhiyuan Liu, Juanzi Li, and Tat-Seng
  Chua. 2019.
\newblock \href {https://www.aclweb.org/anthology/P19-1140} {Multi-channel
  graph neural network for entity alignment}.
\newblock In \emph{Proceedings of the 57th Annual Meeting of the Association
  for Computational Linguistics (ACL)}, pages 1452--1461.

\bibitem[{Chen et~al.(2018{\natexlab{a}})Chen, Dai, Tao, Zhang, Gan, Shen,
  Zhang, Wang, Zhang, and Carin}]{chen2018adversarial}
Liqun Chen, Shuyang Dai, Chenyang Tao, Haichao Zhang, Zhe Gan, Dinghan Shen,
  Yizhe Zhang, Guoyin Wang, Ruiyi Zhang, and Lawrence Carin.
  2018{\natexlab{a}}.
\newblock \href
  {https://proceedings.neurips.cc/paper/2018/hash/074177d3eb6371e32c16c55a3b8f706b-Abstract.html}
  {Adversarial text generation via feature-mover's distance}.
\newblock In \emph{Proceedings of the Annual Conference on Neural Information
  Processing Systems (NeurIPS)}, pages 4671--4682.

\bibitem[{Chen et~al.(2018{\natexlab{b}})Chen, Tian, Chang, Skiena, and
  Zaniolo}]{chen2018co}
Muhao Chen, Yingtao Tian, Kai{-}Wei Chang, Steven Skiena, and Carlo Zaniolo.
  2018{\natexlab{b}}.
\newblock \href {https://doi.org/10.24963/ijcai.2018/556} {Co-training
  embeddings of knowledge graphs and entity descriptions for cross-lingual
  entity alignment}.
\newblock In \emph{Proceedings of the 27th International Joint Conference on
  Artificial Intelligence (IJCAI)}, pages 3998--4004.

\bibitem[{Chen et~al.(2017)Chen, Tian, Yang, and Zaniolo}]{chen_MTransE}
Muhao Chen, Yingtao Tian, Mohan Yang, and Carlo Zaniolo. 2017.
\newblock \href {https://doi.org/10.24963/ijcai.2017/209} {Multilingual
  knowledge graph embeddings for cross-lingual knowledge alignment}.
\newblock In \emph{Proceedings of the 26th International Joint Conference on
  Artificial Intelligence (IJCAI)}, pages 1511--1517.

\bibitem[{Chen et~al.(2020)Chen, Chen, Fan, Uppunda, Sun, and
  Zaniolo}]{chen2020multilingual}
Xuelu Chen, Muhao Chen, Changjun Fan, Ankith Uppunda, Yizhou Sun, and Carlo
  Zaniolo. 2020.
\newblock \href {https://doi.org/10.18653/v1/2020.findings-emnlp.290}
  {Multilingual knowledge graph completion via ensemble knowledge transfer}.
\newblock In \emph{Findings of the Association for Computational Linguistics:
  EMNLP 2020}, pages 3227--3238, Online. Association for Computational
  Linguistics.

\bibitem[{Fey et~al.(2020)Fey, Lenssen, Morris, Masci, and
  Kriege}]{Fey2020Deep}
Matthias Fey, Jan~Eric Lenssen, Christopher Morris, Jonathan Masci, and Nils~M.
  Kriege. 2020.
\newblock \href {https://openreview.net/forum?id=HyeJf1HKvS} {Deep graph
  matching consensus}.
\newblock In \emph{Proceedings of the 8th International Conference on Learning
  Representations (ICLR)}.

\bibitem[{Glorot and Bengio(2010)}]{glorot2010understanding}
Xavier Glorot and Yoshua Bengio. 2010.
\newblock \href {http://proceedings.mlr.press/v9/glorot10a.html} {Understanding
  the difficulty of training deep feedforward neural networks}.
\newblock In \emph{Proceedings of the 13th International Conference on
  Artificial Intelligence and Statistics (AISTATS)}, pages 249--256.

\bibitem[{Goldberger et~al.(2004)Goldberger, Roweis, Hinton, and
  Salakhutdinov}]{goldberger2004neighbourhood}
Jacob Goldberger, Sam~T. Roweis, Geoffrey~E. Hinton, and Ruslan Salakhutdinov.
  2004.
\newblock \href
  {https://proceedings.neurips.cc/paper/2004/hash/42fe880812925e520249e808937738d2-Abstract.html}
  {Neighbourhood components analysis}.
\newblock pages 513--520.

\bibitem[{Hornik et~al.(1989)Hornik, Stinchcombe, and
  White}]{hornik1989multilayer}
Kurt Hornik, Maxwell Stinchcombe, and Halbert White. 1989.
\newblock \href {https://doi.org/10.1016/0893-6080(89)90020-8} {Multilayer
  feedforward networks are universal approximators}.
\newblock \emph{Neural networks}, 2(5):359--366.

\bibitem[{Ji et~al.(2021)Ji, Pan, Cambria, Marttinen, and
  Philip}]{ji2021survey}
Shaoxiong Ji, Shirui Pan, Erik Cambria, Pekka Marttinen, and S~Yu Philip. 2021.
\newblock \href {https://arxiv.org/abs/2002.00388} {A survey on knowledge
  graphs: Representation, acquisition, and applications}.
\newblock \emph{IEEE Transactions on Neural Networks and Learning Systems}.

\bibitem[{Johnson et~al.(2021)Johnson, Douze, and J{\'{e}}gou}]{JDH17}
Jeff Johnson, Matthijs Douze, and Herv{\'{e}} J{\'{e}}gou. 2021.
\newblock \href {https://doi.org/10.1109/TBDATA.2019.2921572} {Billion-scale
  similarity search with gpus}.
\newblock \emph{{IEEE} Transactions on Big Data}, 7(3):535--547.

\bibitem[{Kingma and Ba(2014)}]{kingma2014adam}
Diederik~P Kingma and Jimmy Ba. 2014.
\newblock \href {http://arxiv.org/abs/1412.6980} {Adam: A method for stochastic
  optimization}.
\newblock \emph{arXiv preprint arXiv:1412.6980}.

\bibitem[{Lehmann et~al.(2015)Lehmann, Isele, Jakob, Jentzsch, Kontokostas,
  Mendes, Hellmann, Morsey, van Kleef, Auer, and Bizer}]{lehmann2015dbpedia}
Jens Lehmann, Robert Isele, Max Jakob, Anja Jentzsch, Dimitris Kontokostas,
  Pablo~N. Mendes, Sebastian Hellmann, Mohamed Morsey, Patrick van Kleef,
  S{\"{o}}ren Auer, and Christian Bizer. 2015.
\newblock \href {https://doi.org/10.3233/SW-140134} {{DBpedia} - {A}
  large-scale, multilingual knowledge base extracted from wikipedia}.
\newblock \emph{Semantic Web}, 6(2):167--195.

\bibitem[{Liu et~al.(2021)Liu, Chen, Roth, and Collier}]{liu2021visual}
Fangyu Liu, Muhao Chen, Dan Roth, and Nigel Collier. 2021.
\newblock \href {https://ojs.aaai.org/index.php/AAAI/article/view/16550}
  {Visual pivoting for (unsupervised) entity alignment}.
\newblock In \emph{Proceedings of the 35th {AAAI} Conference on Artificial
  Intelligence (AAAI)}, pages 4257--4266.

\bibitem[{Liu et~al.(2020)Liu, Cao, Pan, Li, Liu, and Chua}]{AttrGNN}
Zhiyuan Liu, Yixin Cao, Liangming Pan, Juanzi Li, Zhiyuan Liu, and Tat{-}Seng
  Chua. 2020.
\newblock \href {https://www.aclweb.org/anthology/2020.emnlp-main.515/}
  {Exploring and evaluating attributes, values, and structures for entity
  alignment}.
\newblock In \emph{Proceedings of the 2020 Conference on Empirical Methods in
  Natural Language Processing (EMNLP)}, pages 6355--6364.

\bibitem[{Pei et~al.(2019)Pei, Yu, and Zhang}]{pei2019improving_OT}
Shichao Pei, Lu~Yu, and Xiangliang Zhang. 2019.
\newblock \href {https://doi.org/10.24963/ijcai.2019/448} {Improving
  cross-lingual entity alignment via optimal transport}.
\newblock In \emph{Proceedings of the 28th International Joint Conference on
  Artificial Intelligence (IJCAI)}, pages 3231--3237.

\bibitem[{Peyr{\'e} et~al.(2019)Peyr{\'e}, Cuturi
  et~al.}]{peyre2019computational_OT}
Gabriel Peyr{\'e}, Marco Cuturi, et~al. 2019.
\newblock \href {http://dx.doi.org/10.1561/2200000073} {Computational optimal
  transport: With applications to data science}.
\newblock \emph{Foundations and Trends{\textregistered} in Machine Learning},
  11(5-6):355--607.

\bibitem[{Radovanovic et~al.(2010)Radovanovic, Nanopoulos, and
  Ivanovic}]{radovanovic2010hubs}
Milos Radovanovic, Alexandros Nanopoulos, and Mirjana Ivanovic. 2010.
\newblock \href {http://portal.acm.org/citation.cfm?id=1953015} {Hubs in space:
  Popular nearest neighbors in high-dimensional data}.
\newblock \emph{Journal of Machine Learning Research}, 11(sept):2487--2531.

\bibitem[{Sun et~al.(2021)Sun, Chen, and Hu}]{sun_no_match}
Zequn Sun, Muhao Chen, and Wei Hu. 2021.
\newblock \href {https://doi.org/10.18653/v1/2021.acl-long.278} {Knowing the
  no-match: Entity alignment with dangling cases}.
\newblock In \emph{Proceedings of the 59th Annual Meeting of the Association
  for Computational Linguistics and the 11th International Joint Conference on
  Natural Language Processing (ACL/IJCNLP)}, pages 3582--3593.

\bibitem[{Sun et~al.(2020{\natexlab{a}})Sun, Chen, Hu, Wang, Dai, and
  Zhang}]{sun-etal-2020-knowledge}
Zequn Sun, Muhao Chen, Wei Hu, Chengming Wang, Jian Dai, and Wei Zhang.
  2020{\natexlab{a}}.
\newblock \href {https://www.aclweb.org/anthology/2020.emnlp-main.460}
  {Knowledge association with hyperbolic knowledge graph embeddings}.
\newblock In \emph{Proceedings of the 2020 Conference on Empirical Methods in
  Natural Language Processing (EMNLP)}, pages 5704--5716.

\bibitem[{Sun et~al.(2017)Sun, Hu, and Li}]{JAPE}
Zequn Sun, Wei Hu, and Chengkai Li. 2017.
\newblock \href {https://doi.org/10.1007/978-3-319-68288-4\_37} {Cross-lingual
  entity alignment via joint attribute-preserving embedding}.
\newblock In \emph{Proceedings of the 16th International Semantic Web
  Conference (ISWC)}, pages 628--644.

\bibitem[{Sun et~al.(2018)Sun, Hu, Zhang, and Qu}]{sun2018bootstrapping}
Zequn Sun, Wei Hu, Qingheng Zhang, and Yuzhong Qu. 2018.
\newblock \href {https://doi.org/10.24963/ijcai.2018/611} {Bootstrapping entity
  alignment with knowledge graph embedding}.
\newblock In \emph{Proceedings of the 27th International Joint Conference on
  Artificial Intelligence (IJCAI)}, pages 4396--4402.

\bibitem[{Sun et~al.(2020{\natexlab{b}})Sun, Wang, Hu, Chen, Dai, Zhang, and
  Qu}]{AliNet}
Zequn Sun, Chengming Wang, Wei Hu, Muhao Chen, Jian Dai, Wei Zhang, and Yuzhong
  Qu. 2020{\natexlab{b}}.
\newblock \href {https://aaai.org/ojs/index.php/AAAI/article/view/5354/5210}
  {Knowledge graph alignment network with gated multi-hop neighborhood
  aggregation}.
\newblock In \emph{Proceedings of the 34th {AAAI} Conference on Artificial
  Intelligence (AAAI)}, pages 222--229.

\bibitem[{Sun et~al.(2020{\natexlab{c}})Sun, Zhang, Hu, Wang, Chen, Akrami, and
  Li}]{sun2020benchmarking}
Zequn Sun, Qingheng Zhang, Wei Hu, Chengming Wang, Muhao Chen, Farahnaz Akrami,
  and Chengkai Li. 2020{\natexlab{c}}.
\newblock \href {http://www.vldb.org/pvldb/vol13/p2326-sun.pdf} {A benchmarking
  study of embedding-based entity alignment for knowledge graphs}.
\newblock \emph{Proceedings of the VLDB Endowment}, 13(11):2326--2340.

\bibitem[{Tang et~al.(2020)Tang, Zhang, Chen, Yang, Chen, and Li}]{BERTINT}
Xiaobin Tang, Jing Zhang, Bo~Chen, Yang Yang, Hong Chen, and Cuiping Li. 2020.
\newblock \href {https://doi.org/10.24963/ijcai.2020/439} {{BERT-INT:} {A}
  bert-based interaction model for knowledge graph alignment}.
\newblock In \emph{Proceedings of the 29th International Joint Conference on
  Artificial Intelligence (IJCAI)}, pages 3174--3180.

\bibitem[{Trisedya et~al.(2019)Trisedya, Qi, and Zhang}]{AttrE}
Bayu~Distiawan Trisedya, Jianzhong Qi, and Rui Zhang. 2019.
\newblock \href {https://doi.org/10.1609/aaai.v33i01.3301297} {Entity alignment
  between knowledge graphs using attribute embeddings}.
\newblock In \emph{Proceedings of the 33rd AAAI Conference on Artificial
  Intelligence (AAAI)}, pages 297--304.

\bibitem[{Wang et~al.(2018)Wang, Lv, Lan, and
  Zhang}]{wang-etal-2018-cross-lingual}
Zhichun Wang, Qingsong Lv, Xiaohan Lan, and Yu~Zhang. 2018.
\newblock \href {https://www.aclweb.org/anthology/D18-1032} {Cross-lingual
  knowledge graph alignment via graph convolutional networks}.
\newblock In \emph{Proceedings of the 2018 Conference on Empirical Methods in
  Natural Language Processing (EMNLP)}, pages 349--357.

\bibitem[{Wang et~al.(2020)Wang, Yang, and Ye}]{WangZhiChun20}
Zhichun Wang, Jinjian Yang, and Xiaoju Ye. 2020.
\newblock \href {https://doi.org/10.18653/v1/2020.emnlp-main.130} {Knowledge
  graph alignment with entity-pair embedding}.
\newblock In \emph{Proceedings of the 2020 Conference on Empirical Methods in
  Natural Language Processing (EMNLP)}, pages 1672--1680.

\bibitem[{Wu et~al.(2019)Wu, Liu, Feng, Wang, Yan, and Zhao}]{RDGCN}
Yuting Wu, Xiao Liu, Yansong Feng, Zheng Wang, Rui Yan, and Dongyan Zhao. 2019.
\newblock \href {https://doi.org/10.24963/ijcai.2019/733} {Relation-aware
  entity alignment for heterogeneous knowledge graphs}.
\newblock In \emph{Proceedings of the 28th International Joint Conference on
  Artificial Intelligence (IJCAI)}, pages 5278--5284.

\bibitem[{Wu et~al.(2020)Wu, Liu, Feng, Wang, and Zhao}]{WuLFWZ20}
Yuting Wu, Xiao Liu, Yansong Feng, Zheng Wang, and Dongyan Zhao. 2020.
\newblock \href {https://doi.org/10.18653/v1/2020.acl-main.578} {Neighborhood
  matching network for entity alignment}.
\newblock In \emph{Proceedings of the 58th Annual Meeting of the Association
  for Computational Linguistics (ACL)}, pages 6477--6487.

\bibitem[{Xu et~al.(2019{\natexlab{a}})Xu, Luo, Zha, and Carin}]{xu2019gromov}
Hongteng Xu, Dixin Luo, Hongyuan Zha, and Lawrence Carin. 2019{\natexlab{a}}.
\newblock \href {http://proceedings.mlr.press/v97/xu19b.html}
  {Gromov-wasserstein learning for graph matching and node embedding}.
\newblock In \emph{Proceedings of the 36th International Conference on Machine
  Learning (ICML)}, pages 6932--6941.

\bibitem[{Xu et~al.(2019{\natexlab{b}})Xu, Wang, Yu, Feng, Song, Wang, and
  Yu}]{GMNN}
Kun Xu, Liwei Wang, Mo~Yu, Yansong Feng, Yan Song, Zhiguo Wang, and Dong Yu.
  2019{\natexlab{b}}.
\newblock \href {https://www.aclweb.org/anthology/P19-1304} {Cross-lingual
  knowledge graph alignment via graph matching neural network}.
\newblock In \emph{Proceedings of the 57th Annual Meeting of the Association
  for Computational Linguistics (ACL)}, pages 3156--3161.

\bibitem[{Yang et~al.(2019)Yang, Zou, Shi, Lu, Lin, and Sun}]{HMAN}
Hsiu-Wei Yang, Yanyan Zou, Peng Shi, Wei Lu, Jimmy Lin, and Xu~Sun. 2019.
\newblock \href {https://www.aclweb.org/anthology/D19-1451} {Aligning
  cross-lingual entities with multi-aspect information}.
\newblock In \emph{Proceedings of the 2019 Conference on Empirical Methods in
  Natural Language Processing and the 9th International Joint Conference on
  Natural Language Processing (EMNLP-IJCNLP)}, pages 4431--4441.

\bibitem[{Zeng et~al.(2021)Zeng, Li, Hou, Li, and Feng}]{zeng2021comprehensive}
Kaisheng Zeng, Chengjiang Li, Lei Hou, Juanzi Li, and Ling Feng. 2021.
\newblock \href
  {https://www.sciencedirect.com/science/article/pii/S2666651021000036} {A
  comprehensive survey of entity alignment for knowledge graphs}.
\newblock \emph{AI Open}, 2:1--13.

\bibitem[{Zhang et~al.(2019)Zhang, Sun, Hu, Chen, Guo, and Qu}]{MultiKE}
Qingheng Zhang, Zequn Sun, Wei Hu, Muhao Chen, Lingbing Guo, and Yuzhong Qu.
  2019.
\newblock \href {https://doi.org/10.24963/ijcai.2019/754} {Multi-view knowledge
  graph embedding for entity alignment}.
\newblock In \emph{Proceedings of the 28th International Joint Conference
  (IJCAI)}, pages 5429--5435.

\end{thebibliography}
\bibliographystyle{acl_natbib}

\appendix

\begin{center}
    {
    \Large\textbf{Appendices}
    }
\end{center}

\section{Dataset Statistics}
We present the dataset statistics of DBP2.0 \citep{sun_no_match} in Tab.~\ref{tab:dataset}. 
DBP2.0 contains three cross-lingual settings for dangling-aware entity alignment, i.e., Chinese-English (ZH-EN), Japanese-English (JA-EN) and French-English (FR-EN). 
Please note that FR-EN is much larger than ZH-EN and JA-EN,
and our methods are scalable to such a large dataset.
\label{sec:appendix}
\begin{table}[!h]	
	\centering
	\resizebox{0.999\columnwidth}{!}{\Large
		\begin{tabular}{clrrcrr}
			\toprule
			\multicolumn{2}{c}{Datasets} & \# Entities & \# Danglings &\# Rel. & \# Triples & \# Align. \\ \midrule
			\multirow{2}{*}{ZH-EN} 
			& ZH & 84,996 & 51,813 & 3,706 & 286,067 &\multirow{2}{*}{33,183} \\
			& EN & 118,996 & 85,813 & 3,402 & 586,868 & \\ \midrule
			\multirow{2}{*}{JA-EN} 
			& JA & 100,860 & 61,090 & 3,243 & 347,204 &\multirow{2}{*}{39,770} \\
			& EN & 139,304 & 99,534 & 3,396 & 668,341 & \\ \midrule
			\multirow{2}{*}{FR-EN}
			& FR & 221,327 & 97,375 & 2,841 & 802,678 &\multirow{2}{*}{123,952} \\
			& EN & 278,411 & 154,459 & 4,598 & 1,287,231 & \\
			\bottomrule
	\end{tabular}}
	\caption{Dataset statistics of DBP2.0}
	\label{tab:dataset}
\end{table}

\section{Computational Environment}\label{appx:computation}
We run experiments on a Linux machine with a single GeForce RTX 2080 Ti GPU with 11 GB GPU memory and a Intel(R) Xeon(R) Gold 6240 CPU @ 2.60GHz. The operating system of our machine is Ubuntu 18.04.2 LTS. The major software packages used are as follows: TensorFlow 1.12; CUDA 10.1; Python 3.6; NumPy 1.18.1; SciPy 1.4.1.
Our source code is available in the attachment for reproducible experiments.

\begin{table*}[!t]
	\centering
	\resizebox{.999\textwidth}{!}
	{\small
		\setlength{\tabcolsep}{3pt}
		\begin{tabular}{lcccccccccccccccccc}
			\toprule
			Methods &
			\multicolumn{3}{c}{ZH-EN} & \multicolumn{3}{c}{EN-ZH} & \multicolumn{3}{c}{JA-EN} & \multicolumn{3}{c}{EN-JA} &  \multicolumn{3}{c}{FR-EN} & \multicolumn{3}{c}{EN-FR}\\
			\cmidrule(lr){2-4} \cmidrule(lr){5-7} \cmidrule(lr){8-10} \cmidrule(lr){11-13} \cmidrule(lr){14-16} \cmidrule(lr){17-19}
			& Prec. & Rec. & F1 & Prec. & Rec. & F1 & Prec. & Rec. & F1 & Prec. & Rec. & F1 & Prec. & Rec. & F1 & Prec. & Rec. & F1 \\ 
			\midrule
			MR & .752 & .538 & .627 & .828 & .505 & .627 & .779 & .580 & .665 & \textbf{.854} & .543 & .664 & .552 & .570 & .561 & \textbf{.686} & .549 & .609 \\
			BR & \textbf{.762} & .556 & .643 & .829 & .515 & .635 & \textbf{.783} & .591 & .673 & .846 & .546 & .663 & .547 & .556 & .552 & .674 & .556 & .609 \\
			\midrule
			\modelname + MR & .750 & .711 & .730 & .838 & \textbf{.726} & \textbf{.778} & .743 & \textbf{.702} & \textbf{.722} & .831 & \textbf{.714} & \textbf{.768 }& .541 & \textbf{.601} & \textbf{.571} & .638 & \textbf{.661} & \textbf{.649} \\
			\modelname + BR & .748 & \textbf{.718} & \textbf{.733} & \textbf{.841} & .721 & .776 & .738 & \textbf{.702} & .719 & .833 & .711 & .767 & \textbf{.556} & .568 & .562 & .681 & .590 & .632 \\
			\bottomrule
	\end{tabular}}
	\caption{Dangling entity detection results on DBP2.0. MR refers to marginal ranking and BR refers to the background ranking~\cite{sun_no_match}. The base alignment model is AliNet~\cite{AliNet}.}
	\label{tab:detection_alinet}
\end{table*}

\begin{table*}[!t]
	\centering
	\resizebox{.999\textwidth}{!}
	{\small
		\setlength{\tabcolsep}{3pt}
		\begin{tabular}{lcccccccccccccccccc}
			\toprule
			Methods &
			\multicolumn{3}{c}{ZH-EN} & \multicolumn{3}{c}{EN-ZH} & \multicolumn{3}{c}{JA-EN} & \multicolumn{3}{c}{EN-JA} &  \multicolumn{3}{c}{FR-EN} & \multicolumn{3}{c}{EN-FR}\\
			\cmidrule(lr){2-4} \cmidrule(lr){5-7} \cmidrule(lr){8-10} \cmidrule(lr){11-13} \cmidrule(lr){14-16} \cmidrule(lr){17-19}
			& Prec. & Rec. & F1 & Prec. & Rec. & F1 & Prec. & Rec. & F1 & Prec. & Rec. & F1 & Prec. & Rec. & F1 & Prec. & Rec. & F1 \\ 
			\midrule
			MR & .207 & \textbf{.299} & .245 & .159 & \textbf{.320} & .213 & .231 & \textbf{.321} & .269 & .178 & \textbf{.340} & .234 & .195 & \textbf{.190} & .193 & .160 & .200 & .178 \\
			BR & .203 & .286 & .238 & .155 & .308 & .207 & .223 & .306 & .258 & .170 & .321 & .222 & .183 & .181 & .182 & .164 & .200 & .180 \\
			\midrule
			\modelname + MR & \textbf{.259} & .280 & \textbf{.269} & .222 & .298 & \textbf{.254} & \textbf{.266} & .288 & \textbf{.276} & \textbf{.225} & .305 & \textbf{.259} & \textbf{.204} & .186 & \textbf{.195} & \textbf{.197} & .189 & \textbf{.193} \\
			\modelname + BR & .258 & .274 & .265 & \textbf{.223} & .305 & .257 & .261 & .281 & .271 & .224 & .306 & .258 & .183 & .180 & .182 & .172 & \textbf{.201} & .185 \\
			\bottomrule
	\end{tabular}}
	\caption{Two-step entity alignment results on DBP2.0. The base alignment model is AliNet.}
	\label{tab:ent_alignment_alinet}
\end{table*}

\begin{table*}[!t]
	\centering
	\resizebox{.999\textwidth}{!}
	{\small
		\setlength{\tabcolsep}{2pt}
		\begin{tabular}{lcccccccccccccccccc}
			\toprule
			\multirow{2}{*}{Methods} &
			\multicolumn{3}{c}{ZH-EN} & \multicolumn{3}{c}{EN-ZH} & \multicolumn{3}{c}{JA-EN} & \multicolumn{3}{c}{EN-JA} & \multicolumn{3}{c}{FR-EN} & \multicolumn{3}{c}{EN-FR}\\
			\cmidrule(lr){2-4} \cmidrule(lr){5-7} \cmidrule(lr){8-10} \cmidrule(lr){11-13} \cmidrule(lr){14-16} \cmidrule(lr){17-19}
			& H@1 & H@10 & MRR & H@1 & H@10 & MRR & H@1 & H@10 & MRR & H@1 & H@10 & MRR & H@1 & H@10 & MRR & H@1 & H@10 & MRR \\ 
			\midrule
			AliNet & .332 & .594 & .421 & {.359} & .629 & .451 & .338 & .596 & .429 & {.363} & .630 & {.455} & .223 & .473 & .306 & .246 & .495 & .329 \\
			\;\;w/ MR & {.343} & {.606} & {.433} & {.364} & {.637} & {.459} & {.349} & {.608} & {.438} & {.377} & {.646} & {.469} & \textbf{.230} & {.477} & \textbf{.312} & {.252} & {.502} & \textbf{.335} \\
			\;\;w/ BR & .333 & .599 & .426 & .357 & .632 & .451 & .341 & {.608} & .431 & .369 & .636 & .461 & .214 & .468 & .298 & .238 & .487 & .321 \\
			\midrule
			\modelname + MR & \textbf{.346} & \textbf{.613} & \textbf{.439} & \textbf{.375} & \textbf{.645} & \textbf{.469} & \textbf{.354} & \textbf{.617 }& \textbf{.444} & \textbf{.379} & \textbf{.654 }& \textbf{.473} & .228 & \textbf{.477} & .311 & \textbf{.253} & .496 & \textbf{.335} \\
			\modelname + BR & .339 & .611 & .432 & .373 & .635 & .464 & .346 & .614 & .437 & .367 & .638 & .460 & .218 & .473 & .303 & .244 & \textbf{.504} & .331 \\
			\bottomrule
	\end{tabular}}
	\caption{Entity alignment results in the relaxed setting on DBP2.0. The base alignment model is AliNet.}
	\label{tab:synthetic_ent_alignment_alinet}
\end{table*}

\section{Hyperparameter Settings}\label{appx:parameters}
To ensure a fair comparison, we follow the hyerparameter settings of the base alignment model (i.e., MTransE and AliNet) and the base dangling detection loss (i.e., MR and BR) reported in the previous work \cite{sun_no_match}.
For our proposed methods, in WGAN, we use a two-layer FNN with 500 hidden units for the critic. 
As suggested by \citet{arjovsky2017wasserstein}, we adopt weight clipping to ensure K-Lipschitz for WGAN and train the critic more than the generator (i.e., the transformation matrix). 
Besides the hyperparameter stated in Section~\ref{sec: experimental_settings}, we tune other hyperparameters within a search space as follows: 
\begin{itemize}
    \item The number of nearest targets $k$: \{5, 10, 15\}
    \item The number of nearest sources $m$: \{5, 10, 15\}
    \item Batch size: \{4096, 8192, 10240, 20480\}
\end{itemize}

\section{More on Experiments}
\label{appendix: more_experiments}
As shown in \citet{sun_no_match}, AliNet \cite{AliNet} performs much worse than MTransE \cite{chen_MTransE} in dangling-aware entity alignment. 
Dangling entity detection would also suffer as a result of the poor alignment performance.
However, in this section, we still present the results of \modelname with AliNet as the base alignment model to  demonstrate that \modelname is model-agnostic and has a good robustness.

\stitle{Consolidated evaluation.} Tab.~\ref{tab:detection_alinet} shows that, using AliNet as the base model, \modelname still outperforms baselines in terms of F1 scores on dangling detection. 
We can see that baselines sometimes achieve better precision with the sacrifice of recall, which leads to unsatisfactory F1 scores.  
Comparing our two variants \modelname + MR and \modelname + BR, there is no one consistently achieving better performance than the other one. 
We report the performance of two-step entity alignment on Tab.~\ref{tab:ent_alignment_alinet}. 
In general, \modelname offers better performance on two-step alignment compared with baselines that do not consider high-order proximities. 
We observe that when we choose AliNet as the base model, the improvement over the baselines is less than the improvement when using MTransE as the base model. 
The reason could be that AliNet generally performs worse than MTransE, even only with MR or BR.
For example, combining Tab.~\ref{tab:detection} and \ref{tab:detection_alinet}, MTransE+MR can achieve 0.740 F1 score, while AliNet+MR only obtains 0.627 F1 score.  
The observation is also pointed out by \citet{sun_no_match}.
The inherent inferiority of AliNet in dangling-aware entity alignment can hinder our new proposed techniques. 
Therefore, we suggest to use MTransE as the base alignment model for dangling-aware entity alignment. 
Future work could investigate other advanced alignment models on this setting. 

\stitle{Relaxed evaluation.}
Tab.~\ref{tab:synthetic_ent_alignment_alinet} demonstrates the results of entity alignment in the relaxed setting.
We observe that AliNet without any dangling detection technique performs the worst. 
By applying dangling detection techniques, the alignment performance increases, indicating that learning to detect dangling entities can indirectly help alignment. 
\modelname with two different base dangling losses (i.e., MR and BR) generally outperforms the corresponding baselines without our proposed techniques. 
For our two variants, \modelname + MR slightly outperforms \modelname + BR variants in most cases.


\section{Computational Cost}
Note that, similar with the MR loss \citep{sun_no_match}, \modelname also relies on nearest neighbor search (NNS) for training. 
Therefore, \modelname can reuse the results of NNS obtained by MR during the training phase, and cause negligible additional overhead. 
On ZH-EN, \modelname averagely spends around 60 seconds training an epoch. 
When the efficiency is of importance in some real-time applications, we can adopt the large-scale efficient similarity search library faiss \citep{JDH17} which uses GPUs for fast NNS. 
Additionally, we could also maintain a cache unit to store the results of NNS and only lazily update the results every ten or twenty epochs during training. 

\section{Limitations} 
We notice that many prior studies on conventional entity alignment consider the side information of entities (e.g., names, descriptions and attributes)~\cite{chen2018co,AttrE,MultiKE,GMNN,WangZhiChun20,WuLFWZ20}. 
However, on dangling-aware entity alignment, the pioneer work \citep{sun_no_match} proposes a framework that only considers the structure information of entities since most KGs are built around relation triples. 
Thus, for a fair comparison, we follow their setting and do not utilize side information of entities. 
Future work could investigate how to effectively incorporate side information for dangling-aware entity alignment in the proper way and with a fair evaluation.

\end{document}